\documentclass{article}

\usepackage{PRIMEarxiv}

\usepackage{enumitem}
\usepackage{comment}
\usepackage{amsmath, amssymb, amsfonts}
\usepackage{hyperref}
\usepackage{url}
\usepackage{booktabs}
\usepackage{nicefrac}
\usepackage{microtype}
\usepackage{lipsum}
\usepackage{fancyhdr}
\usepackage{parskip}
\usepackage{graphicx}
\usepackage{caption}
\usepackage{multirow}
\usepackage{xcolor}
\definecolor{lightgreen}{rgb}{0.56, 0.93, 0.56}
\definecolor{lightred}{rgb}{0.98, 0.81, 0.81}
\definecolor{mygreen}{rgb}{0.0, 0.5, 0.0}
\definecolor{myred}{rgb}{0.9, 0.0, 0.0}
\definecolor{myblue}{rgb}{0.8, 0.8, 1.0}
\usepackage{colortbl}
\usepackage{arydshln}

\graphicspath{{media/}}     
\title{The Confidence-Competence Gap in Large Language Models: A Cognitive Study}

\author{
    \textbf{Aniket Kumar Singh}$^{1,\dag, *}$,
    \textbf{Suman Devkota}$^{2,\dag}$,
    \textbf{Bishal Lamichhane}$^{3,\dag}$,
    \textbf{Uttam Dhakal}$^{2,\dag}$,
    \and \textbf{Chandra Dhakal}$^{4}$
    \\
    $^{1}$Department of Computing and Information Systems, Youngstown State University, Ohio\\
    $^{2}$Electrical and Computer Engineering, Youngstown State University, Ohio\\
    $^{3}$Department of Mathematics and Statistics, University of Nevada, Reno, Nevada\\
    $^{4}$Formerly with the Department of Agricultural and Applied Economics, University of Georgia, Athens, GA 30602\\
$^*$Correspondence: \texttt{aksingh01@student.ysu.edu} (A.K.S.)\\
$^\dag$These authors contributed equally to this work.
}

\begin{document}
\maketitle

\begin{abstract}
    Large Language Models (LLMs) have acquired ubiquitous attention for their performances across diverse domains. Our study here searches through LLMs' cognitive abilities and confidence dynamics. We dive deep into understanding the alignment between their self-assessed confidence and actual performance. We exploit these models with diverse sets of questionnaires and real-world scenarios and extract how LLMs exhibit confidence in their responses. Our findings reveal intriguing instances where models demonstrate high confidence even when they answer incorrectly. This is reminiscent of the Dunning-Kruger effect observed in human psychology.
In contrast, there are cases where models exhibit low confidence with correct answers revealing potential underestimation biases. Our results underscore the need for a deeper understanding of their cognitive processes. By examining the nuances of LLMs' self-assessment mechanism, this investigation provides noteworthy revelations that serve to advance the functionalities and broaden the potential applications of these formidable language models.

\end{abstract}


\section{Introduction}
Ever since the Transformer \cite{Vaswani2017AttentionIA} model was introduced in 2017, we have seen remarkable advancements in the field of Natural Language Processing (NLP) and the recent advent of Large Language Models (LLMs). LLMs have progressed from generating few responses to developing abundant erudite essays. These language models are capable of performing better and learning on their own \cite{Zhao2023ASO}. Large language models have impacted a wide array of fields in a short time. As we all know, the threshold of medical sciences and health is very high. Language models have been proven to be smart enough to cross those barriers \cite{Singhal2023}. These models have been performing different human activities like teaching, organizing business, advertising, being an agent, and content writing. As these models improve and evolve, their behavior becomes increasingly attractive, but at the same time, it is necessary to assess their behaviors from different angles. In recent years, we have seen that these models have emerging capability for attaining human-like intelligence\cite{Wang2023ASO}. Hence, understanding the cognitive abilities of these models is a crucial aspect of responsible and beneficial deployment in real-world scenarios.

Our study is inspired by cognitive science and psychology to investigate the intricacies of LLMs behavior to uncover the mechanism underlying successes and failures at times \cite{Zhuang2023EfficientlyMT} \cite{Shiffrin2023ProbingTP}. Even though these models have showcased their capabilities in generating human-like text, solving complex problems, and reasoning about the world, the mechanism governing their decision-making remains opaque.
As these models are deployed in search engines, writing tools, and other commercial applications, it is essential to understand how these models behave, such as how they think,  the mistakes they make, and how they make decisions \cite{Shiffrin2023ProbingTP}. Adopting innovative evaluation approaches like adaptive testing \cite{Zhuang2023EfficientlyMT} and investigating their capacity for empathy \cite{Huang2023EmotionallyNO}, our study seeks to shed light on the cognitive aspects of LLMs. While we understand these models don't understand things like humans, their skills could change how we think about intelligence. This insight could help intelligence better match what we expect from them in the future. In addition, our study seeks to find if there is a similarity between LLMs behavior and a cognitive phenomenon known as the Dunning-Kruger effect. The Dunning-Kruger effect observed in humans is when people overestimate and underestimate themselves \cite{Kruger1999}.
We carefully inspect the confidence levels revealed by LLMs while they are responding to diverse sets of problems. Even when LLMs don't possess the human capacity of self-awareness, studying their responses and relating them with perceived confidence might offer valuable insight into their self-assessment with correctness. The motivation for this study rises from the fact that as these models get better, it is essential to understand how confident they are in what they do, which will eventually make these models work well in real-life situations.

David Dunning and Justin Kruger conducted several experiments in 1999 \cite{Kruger1999} \cite{Dunning2011}. Dunning and Kruger performed initial research on the phenomenon. Their research finding was very compelling. They highlighted the disconnect between an individual's competence and their perception of competence. 
Our study investigates quantifying self-perceived ability, which is measured through absolute relative confidence levels. This study reveals if a higher confidence level co-relates with higher accuracy. The novelty of our work relies on the fact that we seek the extent of the Dunning-Kruger effect in different LLMs. We dive deep and rigorously into finding out if the models overestimate or underestimate their abilities in specific contexts. Our study reveals appealing perceptions of LLMs' behavior, including situations where models like GPT-4 exhibit high confidence even when their responses are incorrect. This implies a subtle misalignment between self-confidence and self-competence. Likewise, we observed cases where models provided correct answers with shallow confidence, posing queries on underestimation biases. These findings reveal a comparison with the Dunning-Kruger effect. In this well-known cognitive phenomenon, individuals tend to overestimate their abilities in certain domains by clarifying the intricate relationship between cognitive capabilities and levels of confidence in LLMs. This study fosters a deeper understanding of LLMs and their implications for AI applications.
\section{Related Works}
There are several research on large language models. Starting from the approach to information retrieval to the language model replacing human participants \cite{Dillion2023}, the improvement in the capabilities of language models has set an exponential trend. A significant example of advancement in natural language processing is ChatGPT \cite{Baskara2023}. Ouyang et al aligned language models by fine-tuning with a wide range of feedback \cite{Ouyang2022}. Liang and the team presented a holistic evaluation of these models where they validated 25 findings concerning different situations \cite{Liang2022}. Schick et al presented how language models are capable of teaching themselves \cite{Schick2023}. Kraus and the team talk about how language models need to be accurate and integrate their resources to deliver more precise responses \cite{Kraus2023}. 
Yogatama et al analyzed the state of the art of natural language understanding and investigated to evaluate task-independence of the knowledge \cite{Yogatama2019}. They also assessed a metric based on the test data to determine how quickly an existing model can learn new tasks. The study conducted by Acerbi and Stubbersfield examines if LLMs show biases, and they conclude that the presence of biases is widespread in model training data \cite{Acerbi2023}. Our study here focuses on designing the test categories with different levels depending on the questions' complexity. Seven different language models were tested, and their responses were evaluated. 

Drawing inspiration from human cognitive biases,  Erik Jones and J. Steinhardt \cite{Jones2022CapturingFO} study the failures of LLMs, focusing on the need to detect inaccurate behaviors. Hongbin Ye et al.'s study on hallucinations in LLMs \cite{Ye2023CognitiveMA} aligns with our skepticism on LLM-generated outputs, although our work focuses primarily on confidence calibration. They discuss the methods for the detection and improvement of hallucinations by providing a taxonomy of hallucinations. Furthermore, \cite{VeraSorin2023LargeLM} investigated empathy in LLMs, highlighting the significance of social skills. In our paper, we examine the confidence scores(self-assessment scores) before and after the LLMs answer the questions, which aligns with Jiaxin Huang et al.'s work \cite{Huang2022LargeLM}, where they demonstrate the self-improving capabilities of LLMs. Finally, Zhen Lin, Shubhendu Trivedi, and Jimeng Sun's study \cite{Lin2023GeneratingWC} study on uncertainty quantification and the trustworthiness of the models, which relates to our work through confidence estimation. These works highlight the necessity for a thorough understanding of LLM behavior, ranging from cognitive biases and self-improvement to the aspect that our paper focuses on self-assessment and confidence of LLMs.

\newpage 

\section{Methodology}

In this section, we outline our experimental design and procedures for model selection, categorization of test scenarios, and the framework for model interaction. Our goal is to provide a comprehensive overview of the methodology behind our study. For our investigation, we carefully selected a diverse set of large language models (LLMs) to participate in our experiment. These LLMs represent a spectrum of language generation capabilities and are essential for assessing how different models perceive their competence. The selected models include:

\begin{itemize}
    \item GPT-4,  GPT-3.5
    \item BARD,  GooglePaLM 2
    \item LLaMA-2, with three configurations:
    \begin{itemize}
        \item 7 billion parameters
        \item 13 billion parameters
        \item 70 billion parameters
    \end{itemize}
    \item Claude-instant, Claude-2
\end{itemize}

These models were chosen to ensure a comprehensive evaluation of self-assessment abilities across different language generation systems. We employed the native chat interfaces for each model, optimized for their memory and context window capabilities. For open-source models, we leveraged POE.com, a platform by Quora offering access to various LLMs.

\subsection{Test Categories}

Our experiment encompasses a range of distinct test categories, each containing questions of varying complexity. These test categories were carefully crafted to evaluate how LLMs perceive their competence in different knowledge domains. Detailed information on question types, categories, and contexts is provided in Appendix A.

The experiment included the following test categories:

\begin{enumerate}
    \item \textbf{TruthfulQA}: This category featured ten questions spread over five difficulty levels, including Logical Falsehood, Nutrition, Paranormal, Myths and Fairytales, and Fiction.
    
    \item \textbf{TruthfulQA Extended}: Comprising ten questions spread over five difficulty levels, this category included Proverbs, Superstitions, Misquotations, Misconception, and Conspiracies.
    
    \item \textbf{Mathematical Reasoning}: This category covered ten questions, addressing various difficulty levels such as Elementary Mathematics, High School Mathematics, High School Statistics, College Mathematics, and Abstract Algebra.
    
    \item \textbf{LSAT Reasoning}: Consisting of ten questions based on five distinct contexts, each with two associated questions, difficulty escalated from levels 1 to 5.
\end{enumerate}

 The dataset we utilized for this purpose was created with a combination of Benchmarking datasets for LLMs and LSAT Reasoning tests \cite{hendryckstest2021} \cite{hendrycks2021ethics} \cite{zhong2021arlsat} \cite{wang2022lsat}. For a comprehensive understanding of the question types, levels, and contexts, please refer to the Appendix \ref{Surveyquestions}. By structuring our methodology in this way, we aim to provide a detailed and organized account of our experimental procedures, ensuring transparency and rigor in our study.

\subsubsection{Prompt Construction}
In constructing our prompts, we have placed a strong emphasis on maintaining data uniformity and ensuring consistent input structure for each model. To accomplish this objective, we adopted a three-tiered prompting approach. The center of this endeavor is to formulate inquiries in a manner conducive to the comprehension of the language model, thereby mitigating the likelihood of errors resulting from misinterpretation of the posed questions. Our foundational method was the "Simple Prompting technique," a direct and uncomplicated approach that catered to the basic needs of our research. However, for cases where a more nuanced prompting strategy is necessary for a particular model or questions, we have employed the "Chain of Thoughts" (CoT) \cite{Sareen_2023} technique. This method carefully sequences related prompts to foster deeper model engagement and understanding. For those instances where models require even more elaborate and diverse perspectives, we have employed the "Tree of Thoughts" (ToT) \cite{Sareen_2023}  approach. With this technique, we were able to branch out prompts, enabling models to comprehend better and respond to a broader spectrum of related concepts. While our primary goal was to deliver uniform prompts across all models, the integration of the CoT and ToT methods ensured that the distinct needs of specific models were met without undermining the overall consistency of our data.

\subsubsection{Prompt Response Observations}

During our comprehensive evaluation, we presented a standardized set of questions to each language model, meticulously monitoring fluctuations in their confidence levels, all the while refraining from providing any explicit cues pertaining to the complexity of the questions posed. Subsequently, we report the salient findings and performance characteristics of each model. GPT-4 consistently manifested a commendable stability in its confidence levels. Notably, this model displayed an excellent aptitude for processing and generating responses to simple prompts. GPT-3.5 demonstrated adequate prompt comprehension, required minimal prompting, and exhibited increased confidence during the study. Bard maintained a stable confidence level. It exhibited an impressive facility for generating coherent responses to simple prompts without necessitating the deployment of advanced prompting techniques. Google PaLM2 initially displayed well with simple prompts but started generating questions and self-assessing confidence. LLaMA-7B exceeded the performance expectations, showed better prompt comprehension, and rated confidence separately for AC(Absolute Confidence) and RC(Relative Confidence) on individual problems. LLaMA-13B exhibited impressive comprehension speed but struggled with real number questions and showed hesitancy with certain topics. However, it demonstrated perceptible enhancements in response quality when presented with the Chain of Thought (CoT) prompts, along with intermittent reference to prior topics. LLaMA-70B consistently demonstrated a high proficiency in prompt comprehension and, on average, displayed higher levels of confidence in its generated response. Claude-Instant began with lower confidence but gained assurance, emphasizing reliance on training data. Claude-2 responded confidently to simple prompts but struggled with advanced mathematical and LSAT Reasoning, displaying lower confidence and expressing a lack of training for such challenges.

\subsection{Creation of Survey Dataset}

    To rigorously evaluate the performance of Large Language models across various categories and difficulty levels, we have curated an extensive dataset. This dataset not only records the responses generated by the LLMs but also encompasses their self-assessed confidence levels, both before and after their interactions. This offers a clear understanding of the model's intrinsic capabilities and self-awareness. The evaluation of LLMs is determined upon the examination of their diverse answers or responses to the posed questions. Within our dataset, we have incorporated distinct variables that capture the confidence levels of LLMS not only prior to responding to the questions but also subsequent to providing their responses. This inclusive approach enables us to assess the alterations in their confidence levels before and after generating the response. 

Table \ref{tab:dataset_description} in Appendix \ref{variables} provides a detailed description of the variables used in this study. The variables represent the problem's category, difficulty level, their confidence before and after answering the questions, and then the correctness of the response of LLMs. 
In the field of advanced machine learning models, particularly LLMs, evaluating their proficiency goes beyond checking how accurate their output is. It also involves understanding how well these models gauge their abilities, which they express through their confidence levels, and comparing their self-assessment with their actual performance. When we apply these ideas to LLMS, we encounter interesting 
questions. Do LLMs, despite their computations prowess, exhibit similarities to human cognitive biases like the Dunning-Kruger effect? Can we identify the situations where the model is overly confident or lacks confidence in its abilities based on its confidence scores? Our subsequent analyses explore these questions by examining how well the model's self-assessment aligns with its real-world performance. Calibration of confidence levels and their relationship with the accuracy of LLMs are the two significant aspects of our study. These two metrics are examined in the context of the Dunning-Kruger effect. The section oversees the confidence levels and their relation with the accuracy of the LLMs.

\subsection{Introducing Confidence Calibration Metrics} \label{subsec:segmented-analysis}
To determine the calibration of different LLMs based on their reported confidence levels, we segment our data, notably \textit{A1} and \textit{A2}. The following scenarios can be considered:

\begin{enumerate}
    \item \textbf{High Confidence, Correct Answers:} LLMs with high \textit{A1} score (e.g., \( A1 > 7 \)) and correctl answer.
    \item \textbf{High Confidence, Incorrect Answers:} LLMs with high \textit{A1} score but incorrectly answers the question.
    \item \textbf{Low Confidence, Correct Answers:} LLMs with low \textit{A1} score (e.g., \( A1 < 5 \)) and correct answers.
    \item \textbf{Low Confidence, Incorrect Answers:} LLMs with low \textit{A1} score and incorrectly answers the question.

\end{enumerate}
The above information on segmented analysis provides information on how well the confidence level of LLMs are calibrated and how this will relate to the Dunning-Kruger effect. We add a new variable to our dataset that measures the closeness between pre and post-questions confidence scores(\textit{A1} and \textit{A2}, and \textit{R1} and \textit{R2}). Our new variable \textit{Closeness} is defined as:

\[
\text{Closeness} = 
\begin{cases} 
1 & \text{if } |A1 - A2| \leq 1 \\
0 & \text{otherwise}
\end{cases}
\]

We will compare Closeness with \textit{IsCorrect}  to assess if there's any relationship between the LLM's self-assessment accuracy and its performance.

\section{Results}

The data collection process revealed a lot of information about how LLMs behave. In this section, we will discuss the self-assessment abilities of LLMs. Based on the four scenarios created in \ref{subsec:segmented-analysis}, we counted the total number of those instances for each LLM and confidence scores( A\_1, R\_1, etc.). Table \ref{segmentedtable} shows results.

\begin{table}[h]
\centering
\caption{Calibration of Confidence to Competence Across Various Large Language Models (LLMs) for Different Confidence Metrics (A1, A2, R1, R2)}
\label{segmentedtable}
\begin{tabular}{l|l|rrrrrrrrr}
\multirow{2}{*}{\parbox{1.5cm}{Metrics}} & \multirow{2}{*}{} & \multicolumn{9}{c}{Models} \\
 & & \shortstack{Claude-\\2} & \shortstack{Claude-\\Instant} & \shortstack{Google\\Bard} & \shortstack{Google\\PaLM} & \shortstack{GPT-\\3.5} & \shortstack{GPT-\\4} & \shortstack{LLaMA-\\13B} & \shortstack{LLaMA-\\70B} & \shortstack{LLaMA-\\7B} \\
\hline
\multirow{4}{*}{\parbox{2.5cm}{High\_Con\_\\Correct}} & A1 & 3 & 6 & 12 & 1 & 14 & \cellcolor{lightgreen}25 & 5 & 8 & 9 \\
 & A2 & 14 & 13 & 18 & 8 & 21 & \cellcolor{lightgreen}25 & 8 & 14 & 8 \\
 & R1 & 3 & 0 & 21 & 0 & 12 & \cellcolor{lightgreen}25 & 5 & 8 & 5 \\
 & R2 & 13 & 3 & 21 & 4 & 21 & \cellcolor{lightgreen}25 & 5 & 13 & 9 \\
\hline
\multirow{4}{*}{\parbox{2.5cm}{High\_Con\_\\Incorrect}} & A1 & 3 & 2 & 6 & 1 & 5 & 15 & \cellcolor{lightgreen}23 & 18 & 6 \\
 & A2 & 4 & 21 & 14 & 6 & 16 & 15 & \cellcolor{lightgreen}25 & 22 & 21 \\
 & R1 & 3 & 0 & \cellcolor{lightgreen}17 & 2 & 5 & 15 & 13 & 18 & 6 \\
 & R2 & 4 & 7 & 15 & 2 & 16 & 15 & 16 & \cellcolor{lightgreen}22 & 14 \\
\hline
\multirow{4}{*}{\parbox{2.5cm}{Low\_Con\_\\Correct}} & A1 & 2 & 0 & 0 & \cellcolor{lightgreen} 3 & 0 & 0 & 0 & 0 & 0 \\
 & A2 & 0 & 0 & 0 & 0 & 0 & 0 & 0 & 0 & 0 \\
 & R1 & 2 & 1 & 0 & \cellcolor{lightgreen}6 & 0 & 0 & 1 & 0 & 2 \\
 & R2 & 0 & 0 & 0 & 0 & 0 & 0 & 0 & 0 & 0 \\
\hline
\multirow{4}{*}{\parbox{2.5cm}{Low\_Con\_\\Incorrect}} & A1 & \cellcolor{lightgreen}12 & 2 & 0 & 5 & 0 & 0 & 2 & 0 & 2 \\
 & A2 & \cellcolor{lightgreen}14 & 0 & 0 & 0 & 0 & 0 & 2 & 0 & 0 \\
 & R1 & 12 & 5 & 0 & \cellcolor{lightgreen}16 & 0 & 0 & 3 & 0 & 0 \\
 & R2 & \cellcolor{lightgreen}14 & 0 & 0 & 6 & 0 & 0 & 2 & 0 & 0 \\
\end{tabular}
\end{table}

From Table \ref{segmentedtable}, we can see that models like GPT-4 show a high number of correct answers when confident (High\_Conf\_Correct\_A1 = 25). But if we look at the High\_Conf\_Incorrect scores, it is 15. While this score is not the highest compared to other models, it is high, and this means GPT-4 was always highly confident in itself while answering the questions we provided(regardless of the correctness). LLaMA-13B also shows a discrepancy in high confidence and actual performance, with High\_Conf\_Incorrect\_A1 at 23 instances. This could be interpreted as a potential misalignment between confidence and competence, akin to the overestimation seen in the Dunning-Kruger effect. Claude-Instant has High\_Con\_Incorrect\_A2 of 21. This means more than half of the time, Claude-Instant was highly confident after answering the question but got it incorrect. Google-PaLM, with a Low\_Conf\_Correct\_A1 of 3, shows cases where the model is correct despite low confidence. While it is not conclusive, this could be a point of investigation for underestimation biases. 
 Google-Bard shows similar High\_Conf\_Correct and High\_Conf\_Incorrect scores before (A1) and after answering (A2), suggesting a more stable confidence calibration similar to GPT-4. Actually, Google-Bard is also overconfident( high High\_Con\_Incorrect scores), similar to GPT-3.5 and GPT-4.

The evidence from our result is a strong inclination toward cognitive biases like the Dunning-Kruger effect in LLMs.  While we must exercise caution before jumping to any conclusion, our data contains scenarios where LLMs' high confidence does not always correlate with correct answers and vice versa. However, these are hands-on evidence, and we don't recommend it to be definitive evidence of such psychological phenomena in these models. Our study is a practical test to determine how well these models behave. It helps us try to understand why these models sometimes act overconfident or provide incorrect information because of the way they process information or what we see because of the data they are exposed to. 

\subsection{Confidence Closeness }

In the section above, we looked at how the correctness of LLM is compared to their confidence. To take this one step further, we will look at their correctness when based on the variable created in section \ref{subsec:segmented-analysis}. The relation between 'A1' and 'A2' on how close they are pre-task and post-task serves as an indicator for LLMs on how consistent the self-assessment is.
 A high value in the "Close\_Correct" category implies that the model is generally correct while being confident. In addition to that, it is also an indication that the model maintains a consistent level of confidence before and after answering the question. On the other side, a high count in the "Close\_Incorrect" category suggests that the models' confidence is stable even if their answers are incorrect. The results are summarized in Table \ref{closeness} in Appendix \ref{apcolvscor}.

\begin{figure}[h!]
    \centering
    \includegraphics[width=0.7\textwidth]{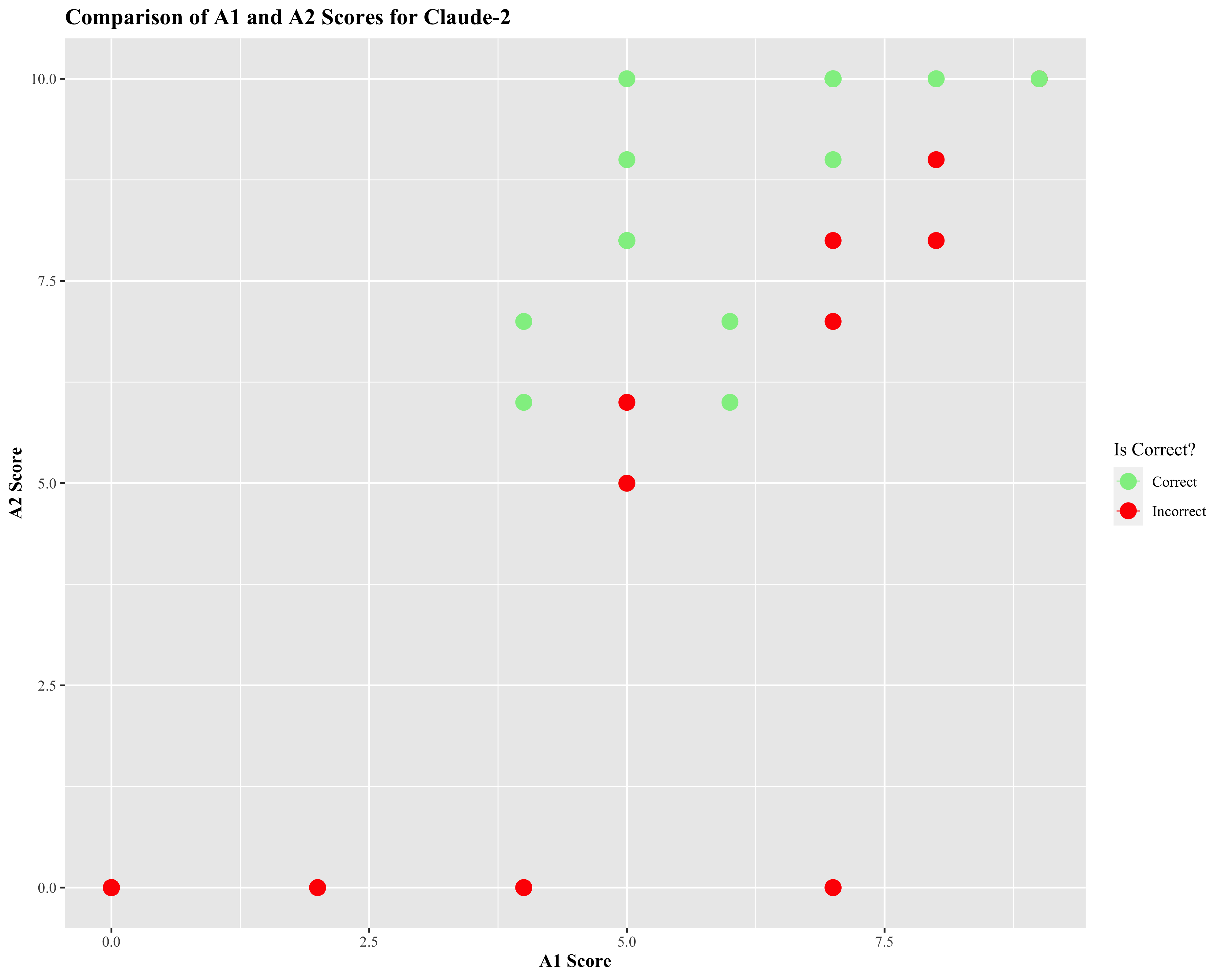}
    \caption{Comparison of A1 and A2 Scores for Claude-2.}
    \label{fig:claude2plot}
\end{figure}

As we have seen above, GPT-4 was very confident in its response regardless of the correctness of the answer. We can see a similar pattern in this case, too. Claude-2 shows a lower "Close\_Correct" but a higher "Close\_InCorrect" and "Far\_Correct" count. This is evidence that Claude-2 is not able to evaluate itself, as when the confidence score was closer to each other, it had 14 incorrect responses out of 40 responses. Still, when the confidence scores were far from each other, it had 15 correct out of 40. This suggests two things:1) either Claude-2 initially had a low A1. After answering the question, it increased its confidence score, A2, and then got it correct, or 2) it initially had a high A1 but later lowered its confidence, but it still got it right. The first one tells us that Claude-2 was able to change and update its evaluation correctly. Figure \ref{fig:claude2plot} illustrates Claude-2's confidence score to reflect their evaluating behavior. The four red dots on the x-axis tell us that Claude-2 successfully lowered its confidence score after answering the question, and the answer was incorrect. This means Claude-2 was able to successfully assess itself after looking at the question for these four instances.
In most cases (shown by the green dots), when it increased its confidence after looking at the question, it got the answers correct. However, it did increase the confidence but yet got incorrect answers in some cases. A similar observation was found for LLaMA-13B, where it has high counts in "Close\_InCorrect." The zero count for GoogleBard in Far\_Correct and Far\_Incorrect tells us that its evaluation is pretty much the same before and after answering the question. Table \ref{closeness} in Appendix \ref{apcolvscor} shows the complete result for all LLMs. 

\subsection{Distribution of Confidence Scores}

\begin{figure}[htbp]
  \centering
  \includegraphics[width=0.8\textwidth]{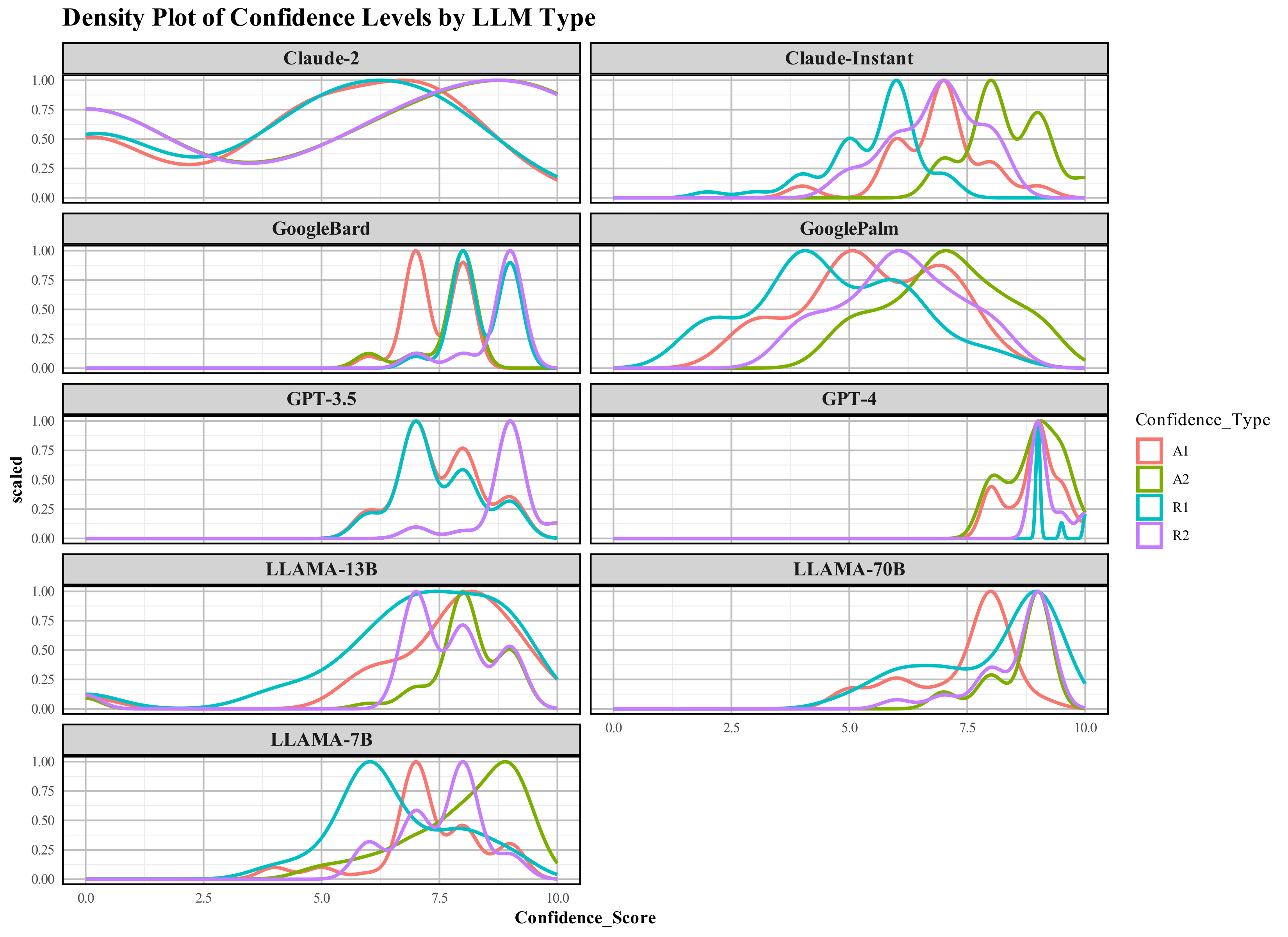}
  \caption{Facetted Density Plot of Confidence Levels by LLM Type. The plot reveals varying patterns of confidence distribution across different LLM types, suggesting nuanced self-perceptions in these models.}
  \label{fig:facet_density_plot}
\end{figure}

\begin{table}[h]
\centering
\caption{Summary Statistics of Confidence Scores by LLM Type}
\label{table:summary_stats}
\begin{tabular}{lcccccccc}
\hline
LLM & \multicolumn{2}{c}{A1} & \multicolumn{2}{c}{R1} & \multicolumn{2}{c}{A2} & \multicolumn{2}{c}{R2} \\
    & Mean & SD & Mean & SD & Mean & SD & Mean & SD \\
\hline
Claude-2 & 4.800 & 2.911 & 4.650 & 2.957 & 5.400 & 4.241 & 5.400 & 4.235 \\
Claude-Instant & 6.850 & 1.027 & 5.475 & 1.062 & 8.325 & 0.829 & 6.825 & 0.931 \\
GoogleBard & 7.400 & 0.591 & 8.400 & 0.591 & 7.700 & 0.648 & 8.700 & 0.648 \\
GooglePaLM & 5.500 & 1.485 & 4.600 & 1.646 & 7.050 & 1.260 & 6.050 & 1.260 \\
GPT-3.5 & 7.525 & 0.877 & 7.475 & 0.877 & 8.900 & 0.672 & 8.900 & 0.672 \\
GPT-4 & 8.900 & 0.568 & 9.200 & 0.372 & 8.925 & 0.594 & 9.225 & 0.375 \\
LLaMA-13B & 7.550 & 2.062 & 6.950 & 2.136 & 7.725 & 1.921 & 7.400 & 1.892 \\
LLaMA-70B & 7.350 & 1.122 & 7.950 & 1.339 & 8.600 & 0.672 & 8.475 & 0.847 \\
LLaMA-7B & 7.250 & 1.214 & 6.600 & 1.297 & 8.025 & 1.187 & 7.525 & 0.877 \\
\hline
\end{tabular}
\end{table}

The facetted destiny plot in Figure \ref{fig:facet_density_plot}  with the summary of statistics given in table \ref{table:summary_stats} presents the distinct patterns in self-assessment across different LLMs.
The mean confidence level for A1 and R1 of Claude-2 is approximately 4.8 and 4.65, respectively. These mean confidence levels are coupled with higher standard deviations of 2.91 and 2.95 simultaneously. The high standard deviation for confidence level directs toward a broad spectrum of self-perceived abilities. In addition, the post-task mean confidence level for A2 and R2 is also higher, with a higher standard deviation. Higher Standard deviation for A2 and R2 implies significant inconsistencies in self-assessment after completion of the task.
Individually, the mean confidence score of A1 and R1 for Claude-Instant is 6.85 and 5.47, respectively, with a lower standard deviation of 1.03 and 1.06 simultaneously. The confidence after completing the task spiked to 8.32 and 6.82 for A2 and R2, maintaining the low variability of data around 0.83 and 0.93, respectively. \\
Even though Google-Bard generally outperforms Google-PaLM across the board, both of these models maintained consistent confidence metrics. In addition, model GPT-3.5 and 4 also encompasses high mean confidence levels. GPT-4 shows a mean A1 confidence score of 8.9 with a standard deviation of 0.568.
Among the LLaMA series, variability in confidence levels is more noticeable.
LLaMA-13B has a standard deviation of 2.06 for A1, which is higher, While series LLaMA-70B and LLaMA-7B are in the range of 1.12 and 1.21, respectively.
To summarize, the findings here are detailed in the self-assessed confidence levels with various LLMs. The destiny plot in upcoming sections will further
illustrate the trends, where the curve varies in width and height, implying the observed mean and variability in confidence levels. These results underscore the fact that our analysis should consider both central tendency and dispersion for self-assessment mechanisms of LLMs.

\begin{figure}[htbp]
    \centering
    \includegraphics[width=\textwidth]{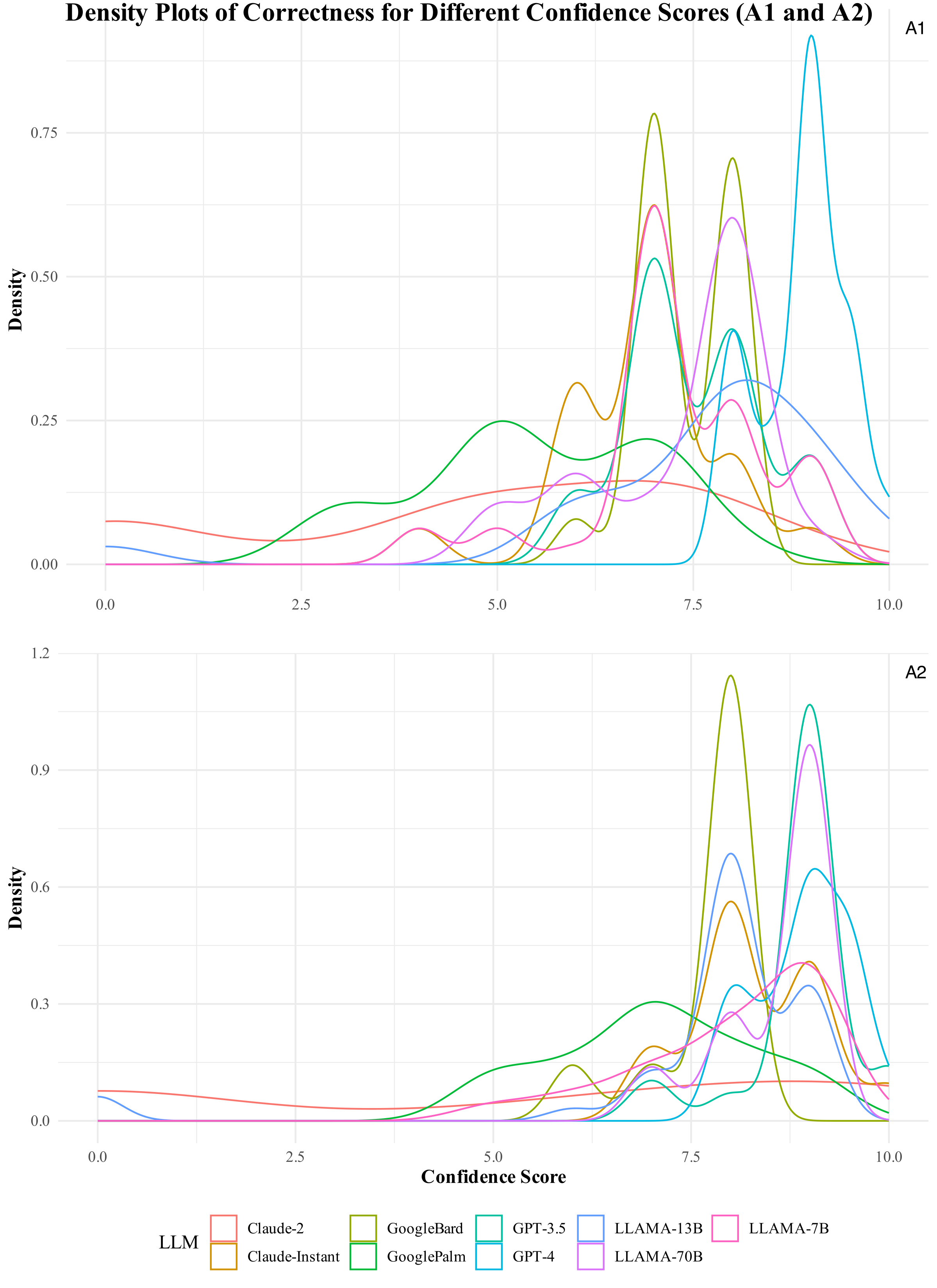}
    \caption{Density Plots of Correctness for Different Confidence Scores (A1 and A2)}
    \label{fig:A1A2}
\end{figure}

The density plot illustrated in Figure~\ref{fig:A1A2} shows the distribution of confidence scores across different LLMs for both A1 and A2 scores. A similar distribution plot for R1 and R2 is in Appendix \ref{Distcon} Figure \ref{fig:R1R2}. We can compare the distributions across different LLMs and observe how their confidence scores vary. For instance, the density plot for A1 in Figure \ref{fig:A1A2} shows us that GPT-4 is very confident in most of the cases. Figure \ref{fig:facet_density_plot}, \ref{fig:A1A2}, and  \ref{fig:R1R2} give us an initial picture of the variation of confidence scores in LLMs. Now we will incorporate the correctness in the mix to study how well these LLMs do. 

\newpage
\subsection{Category vs Confidence Scores}

Understanding how LLMs self-assess their performance via confidence can provide valuable perception of their limitations and capabilities.
Our data set provides a significant variation of LLMs performance across several categories like LSAT- Reasoning, Mathematical Reasoning, and Truthful Q\&A. Our data set portrays confidence scores both before (A1 and R1) and after (A2 and R2) answering the questions in above mentioned categories.

With reference to confidence, GPT-4 was able to succeed in setting itself apart from others with consistency in its high absolute and relative pre-task and post-task confidence levels across all tested categories.
Significantly, it exhibited unparalleled confidence in the LSAT Reasoning task. This implies that it has more vital abilities in logical and analytical Reasoning.
In contrast, Claude-2 and Claude-Instant represented a less consistent confidence profile. Even though Claude-2 demonstrated diminished pre-task and post-task confidence levels in LSAT Reasoning, its confidence tends to improve in the truthful Q\&A category. The variation of this confidence advocates Claude-2 and similar models may be optimized for specific types of tasks, ergo influencing their self-assessed confidence. For a detailed review, readers are encouraged to refer to Appendix Table \ref{CatVsCon}.  The significant differences in how confident the models are could help us understand how well they work for different types of problems. When we observe the apparent differences in confidence among the models, it can provide us with valuable insights into how well they can be applied to various types of problems.

\begin{figure}[h!]
    \centering
    \includegraphics[width=\textwidth]{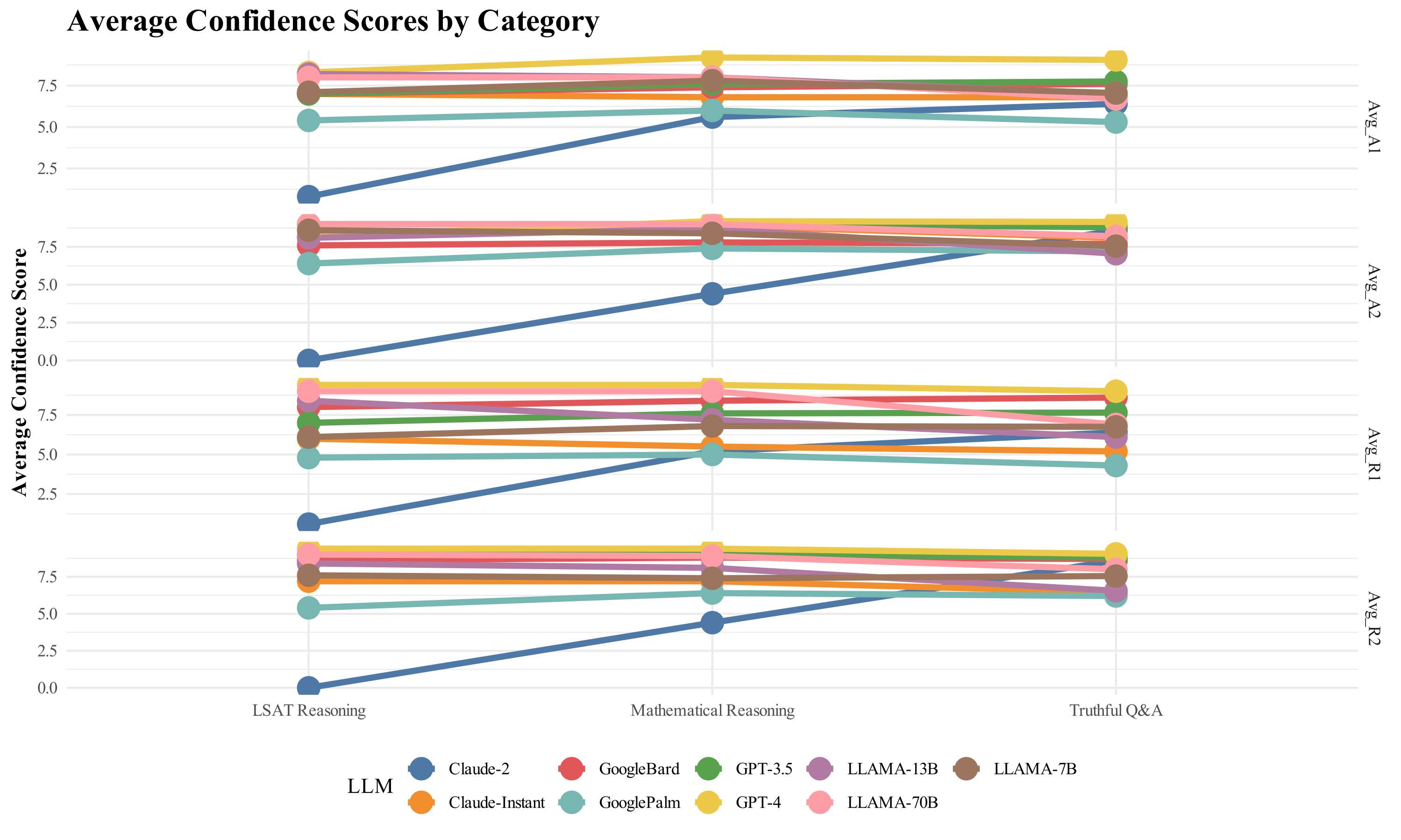}
    \caption{Average Confidence Levels by Category and LLM}
    \label{fig:CatVsCon_plot}
\end{figure}
In addition, models like LLaMA-70B shows high confidence score for LSAT reasoning and Mathematical Reasoning; however, they possess lower confidence score in the Truthful Q\&A category. This kind of variability across different types suggests models might have nuanced areas of expertise.
Such within-model variability across different categories suggests that individual models may have nuanced areas of expertise. This tells us that we should be smart enough to pick a model based on the categories.

It is noteworthy to mention the anomaly observed with Claude-2 in LSAT Reasoning, where it recorded an extremely low confidence level, particularly for the post-task metrics (A2 and R2). While the reason for this remains elusive, it raises questions about the model's internal evaluation mechanisms or possible computational errors that need further careful observation.
Our analysis uncovers the complex landscape of confidence with different LLMs and the problem categories that they are prompted to. A model like GPT-4 appeared to be general for all the various tasks maintaining high confidence. However, other LLMs seem to be specialized in a specific domain or yet to be tuned for generalization. Our findings convey the information to consider model-specific confidence evaluation while selecting a particular model for a specific task. This also opens new interest in the research area for LLMs to understand problem difficulty and correctness of the answer(IScorrect), offering more perspective on performance and self-assessment of the model. As seen in Figure~\ref{fig:CatVsCon_plot}, there is a noticeable pattern in the confidence levels across different problem categories and LLMs.

Our data set is based on comprehensive performance metrics, and the plots are generated utilizing these datasets. Conspicuously, models like GPT-4 and LLaMA-70B hold higher post-task confidence levels across all examined categories. Mathematical Reasoning seems to be standing out with consistency in high confidence levels, recommending that models are more secure in their performance in mathematical tasks in comparison to other types of functions. Our experimental data on the 'Truthful Q\&A' category displays variable performance, suggesting that the nature of a task might affect LLMs' confidence distinctively. These variations in confidence levels should be considered to have a practical implication for the development of LLMs in specializing in particular tasks.

\subsection{Problem Level vs. Confidence Scores}

\begin{figure}[h!]
    \centering
    \includegraphics[width=\textwidth]{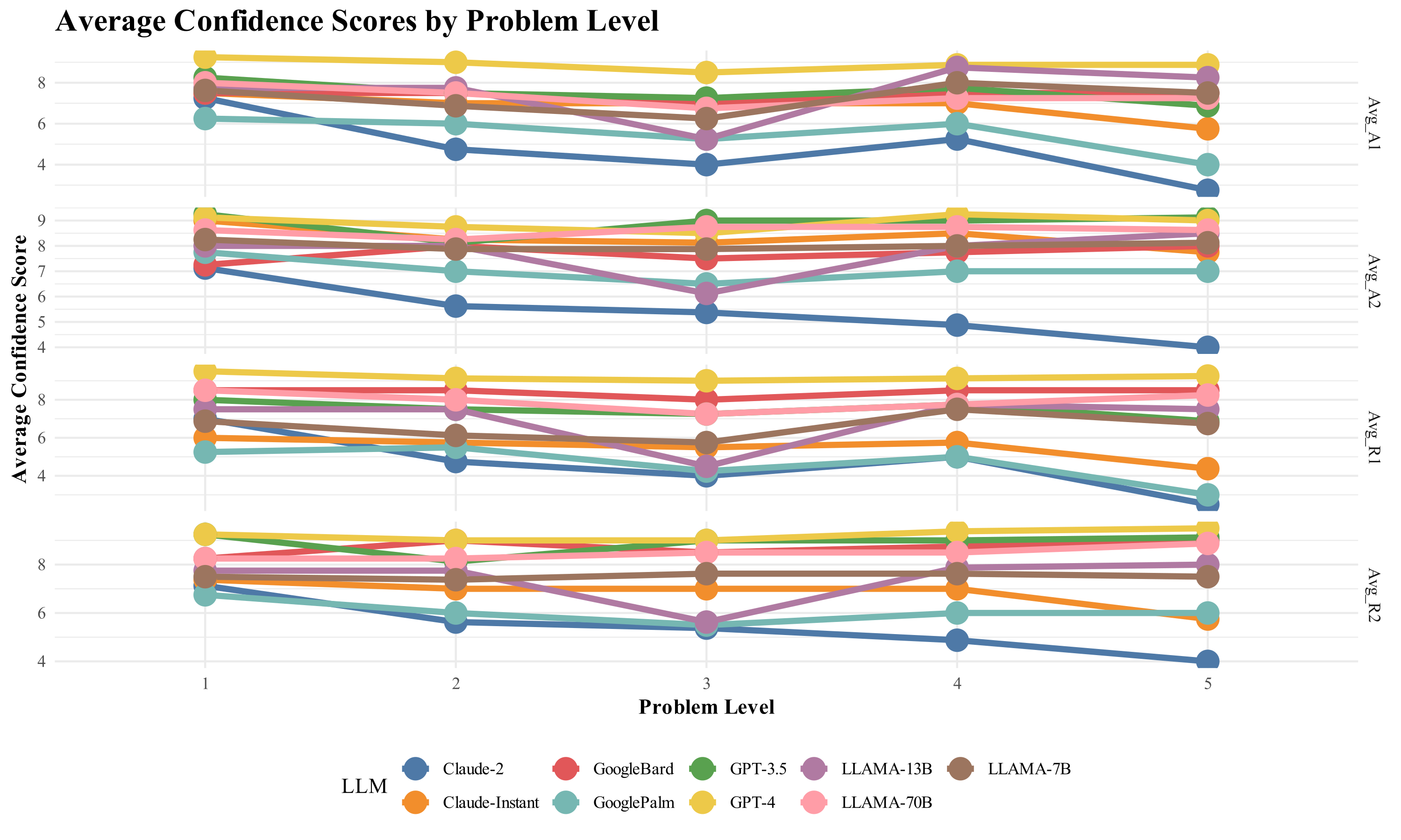}
    \caption{Average Confidence Scores by Problem Level}
    \label{fig:prob_level_plot}
\end{figure}

The dataset table in Appendix \ref{prolevVsCon} Table \ref{tab:problevelvscon} illuminates the average confidence scores (both absolute and relative) expressed by different LLMs at different problem levels (ranging from 1 to 5). The visualization for the table is represented in Figure \ref{fig:prob_level_plot}. Predominantly, it is conspicuous that as the level of problem increases, the confidence score for LLMs decreases. The pattern is very noticeable with the absolute confidence score. LLMs felt less sure about their answers as the level of the problem increased. This result advocates that LLMs may struggle to sustain high confidence when prompted with convoluted tasks.

In contrast, a relative confidence score doesn't follow this trend. Even though there is a slight reduction in relative confidence as the level of problem is increased, it is not as steep as the drop in absolute confidence.
This implies that LLMs might understand their performance in comparison to others as relatively stable across different problem levels.

In addition, it is worth acknowledging that each LLM differs in their response to problem level. To illustrate, GPT-4 maintained a high confidence score across all problem levels, indicating its consistency in self-assessment of its performance. However, models like Claude-2 and Claude-Instant represented higher variability in their confidence scores as the level of the problem changed. This is another indication that some models may adapt differently to task difficulties. For example, Claude-2 shows a notable improvement in confidence levels for certain problems.  

It is imperative to underscore that our analysis provides an overview of the average trends in confidence scores across different problem levels. Individual variations in model behaviors are manifested, and this observation showcases the need for a nuanced understanding of how different language models respond to diverse problem complexities. In addition to this, further investigation with the incorporation of additional factors like problem content and correctness of answer could offer valuable insight into LLMs performance and confidence assessment in particular scenarios.

\section{Discussion \& Conclusion}

In this study, we analyzed the self-assessment behavior of Large language models such as GPT, Claude, LLaMA, and others through their confidence scores, investigating potential parallels with the Dunning-Kruger effect. 
As depicted in table \ref{CatVsCon} and figure \ref{fig:CatVsCon_plot} provides intriguing insights about how LLMs assess their performance across different categories. Even though our study didn't establish a solid presence of the Dunning-Kruger effect, it provided valuable observations aligning with its conceptual framework. 

GPT-4 stands out with its consistency in high confidence scores across all the tested categories, especially in LSAT Reasoning tasks. Such a high confidence pattern resembles its high ability to gauge its competence accurately. Nevertheless, it is essential to be careful to jump to any conclusion as there might be other factors contributing to this trend. On the other hand, models like Claude-2 and Claude Instant displayed higher variability in their confidence scores across different categories. Claude-2 showed a relatively low confidence score for LSAT Reasoning; however, they performed better in Truthful Q\&A. This difference mirrors the concept of individuals with varying abilities showing inconsistency in assessments. For now, this observation serves as a parallel instead of conclusive proof of the Dunning-Kruger effect's applicability in this context. LLaMA-70B performed better with a higher confidence score in LSAT Reasoning and Mathematical categories but had lower confidence in Truthful Q\&A. This subtle variation aligns with the idea that individual LLMs might possess specialized domains of competence, akin to the Dunning-Kruger effect's recognition of skill variations among individuals.

Referencing Table \ref{tab:problevelvscon} and Figure \ref{fig:prob_level_plot}, we investigate and explore the relationship between problem-level complexity and LLM confidence scores. The observed patterns in confidence offer exciting and interesting connections to the Dunning-Kruger effect, even if they don't provide solid evidence of it. LLMs were observed to possess higher confidence scores starting at level 1. With increasing complexity, a decrease in confidence score was observed.
The observed overconfidence phase is related to the overestimation part of the Dunning-Kruger effect, wherein individuals with lower abilities often overrate their competence.  Different LLMs exhibited varying confidence score patterns across the problem levels, reflecting the notion that individuals with different abilities experience varying degrees of the Dunning-Kruger effect. Also, Models like GPT-4 maintained their confidence similar to the individual with high abilities making accurate self-assessments.

In a nutshell, the pattern of the LLM's confidence score provides intriguing parallels with the Dunning-Kruger effect. However, they don't provide solid evidence of its presence in the behavior of LLMs. To provide a sturdy connection, further research with statistical analysis and a broader set of variables is vital. Our findings, nevertheless, pave the way for deeper exploration of LLMs with the Dunning-Kruger effect by showing the relationship between self-assessment and competence in artificial intelligence. The underlying intricacies of LLM behavior, the biases, and the confidence framework demand a more in-depth, comprehensive exploration. It opens doors to a myriad of questions that deserve attention, hinting at a treasure trove of insights awaiting to dive deep. Delving deeper into this convergence of psychology and artificial intelligence offers a promising frontier, potentially unlocking novel insights into AI behavior and ethics. The observations from this study beckon a broader exploration, suggesting that the mysteries of AI cognition, akin to human nuances, are both vast and awaiting discovery.

\newpage

\bibliographystyle{unsrt}  
\bibliography{references}  

\newpage

\appendix
\section{Data}
\subsection{Survey Questions } \label{Surveyquestions}

\begin{enumerate}
    \item \textbf{TruthfulQA}: 
    Included ten questions spread over five difficulty levels, with two questions per level. The levels were:
    \begin{enumerate}[label=\textbf{Level \arabic*:}, leftmargin = 2.5cm] 
        \item Logical Falsehood
        \item Nutrition
        \item Paranormal
        \item Myths and Fairytales
        \item Fiction
    \end{enumerate}
    
    \item \textbf{TruthfulQA Extended}:
    Ten questions spread over five difficulty levels, two per level. The levels were:
    \begin{enumerate}[label=\textbf{Level \arabic*:}, leftmargin = 2.5cm]
        \item Proverbs
        \item Superstitions
        \item Misquotations
        \item Misconception
        \item Conspiracies
    \end{enumerate}
    
    \item \textbf{Mathematical Reasoning}:
    Spanning ten questions across:
    \begin{enumerate}[label=\textbf{Level \arabic*:},  leftmargin = 2.5cm]
        \item Elementary Mathematics
        \item High School Mathematics
        \item High School Statistics
        \item College Mathematics
        \item Abstract Algebra
    \end{enumerate}
    
    \item \textbf{LSAT Reasoning}:
    Comprising ten questions based on five distinct contexts. Each context had two associated questions, with difficulty escalating from levels 1 to 5.
\end{enumerate}

\subsection{Variables} \label{variables}

\begin{table}[h]
    \centering
    \caption{Description of the Dataset Variables}
    \label{tab:dataset_description}
    \begin{tabular}{l l l l}
    \toprule
    \textbf{Variable Symbol} & \textbf{Variable Name} & \textbf{Type} & \textbf{Range/Example} \\
    \midrule
    Category      & Category of the problem        & Categorical & Truthful Q\&A, Mathematical Reasoning \\
    ProblemLevel  & Problem level                  & Categorical & 1, 2, 3, 4, 5 \\
    ProblemID     & Unique identifier for problem  & Categorical & T1, T2, MR1 \\
    Problem       & Text of the problem            & Text        & \textit{"Are all real numbers real numbers?" } \\
    LLM           & Type of Large Language Model   & Categorical & GPT-4, GPT-3.5 \\
    A1            & Absolute Confidence (Pre)      & Continuous  & 1-10 \\
    R1            & Relative Confidence (Pre)      & Continuous  & 1-10 \\
    A2            & Absolute Confidence (Post)     & Continuous  & 1-10 \\
    R2            & Relative Confidence (Post)     & Continuous  & 1-10 \\
    IsCorrect     & Correctness of answer          & Binary      & 0, 1 \\
    \bottomrule
    \end{tabular}
\end{table}
\textbf{A1}: Absolute confidence level expressed by the LLM before answering. Question: "How well do you think you will do?" \\
\textbf{R1}: Relative confidence level expressed by the LLM before answering, compared to others. Question: "Compared to others, how well do you think you will do?" \\ 
\textbf{A2}: Absolute confidence level expressed by the LLM after answering. Question: "How well do you think you did?" \\ 
\textbf{R2}:  Relative confidence level expressed by the LLM after answering, compared to others. Question: "Compared to others, how well do you think you did?"

\section{Tables and Figures}
\subsection{Closeness vs correctness} \label{apcolvscor}
\begin{table}[h]
\centering
\caption{Your Caption Here}
\label{closeness}

\begin{tabular}{l|rrrr}
  \multicolumn{5}{c}{\textbf{Absolute Confidence}} \\
  \textbf{LLM} & \textbf{Close\_Correct} & \textbf{Close\_Incorrect} & \textbf{Far\_Correct} & \textbf{Far\_Incorrect} \\
  \hline
  Claude-2 & \cellcolor{lightred}4 & 15 &  \cellcolor{lightgreen}15 & 6 \\
  Claude-Instant & 11 & 12 & 5 & 12 \\
  GoogleBard &22  & 18 & \cellcolor{lightred}0 & \cellcolor{lightred}0 \\
  GooglePaLM & 9 & 9 & 7 & \cellcolor{lightgreen}15 \\
  GPT-3.5 & 14 & \cellcolor{lightred}5 & 9 & 12 \\
  GPT-4 & \cellcolor{lightgreen}25 & 15 & \cellcolor{lightred}0 & \cellcolor{lightred}0 \\
  LLaMA-13B & 7 & \cellcolor{lightgreen}24 & 2 & 7 \\
  LLaMA-70B & 8 & 20 & 8 & 4 \\
  LLaMA-7B & 9 & 14 & 6 & 11 \\
\end{tabular}
\quad 
\begin{tabular}{l|rrrr}
  \multicolumn{5}{c}{\textbf{Relative Confidence}} \\
  \textbf{LLM} & \textbf{Close\_Correct} & \textbf{Close\_Incorrect} & \textbf{Far\_Correct} & \textbf{Far\_Incorrect} \\
  \hline
  Claude-2 & \cellcolor{lightred}4 & 16 & \cellcolor{lightgreen}15 & 5 \\
  Claude-Instant & 12 & 13 & 4 & 11 \\
  GoogleBard & 22 & 18 & \cellcolor{lightred}0 & \cellcolor{lightred}0 \\
  GooglePaLM & 9 & 9 & 7 & \cellcolor{lightgreen}15 \\
  GPT-3.5 & 14 & \cellcolor{lightred}5 & 9 & 12 \\
  GPT-4 & \cellcolor{lightgreen}25 & 15 & \cellcolor{lightred}0 & \cellcolor{lightred}0 \\
  LLaMA-13B & 7 & \cellcolor{lightgreen}26 & 2 & 5 \\
  LLaMA-70B & 10 & 20 & 6 & 4 \\
  LLaMA-7B & 8 & 15 & 7 & 10 \\
\end{tabular}

\end{table}

\newpage
\subsection{Distribution of Confidence Scores} \label{Distcon}

\begin{figure}[htbp]
    \centering
    \includegraphics[width=0.8\textwidth]{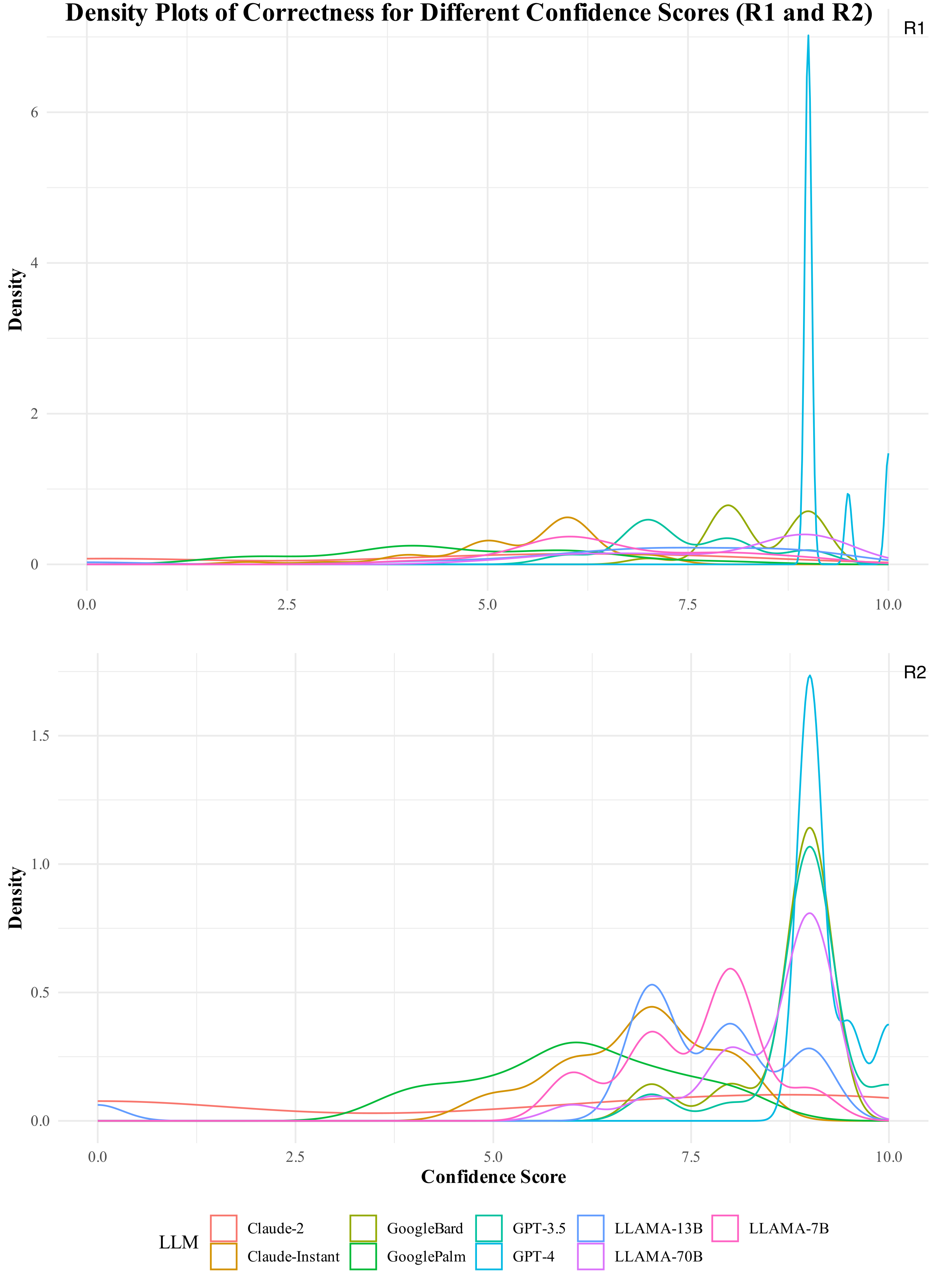}
    \caption{Density Plots of Correctness for Different Confidence Scores (R1 and R2)}
    \label{fig:R1R2}
\end{figure}

\newpage 

\subsection{Confidence Scores vs Correctness} \label{CorvsCon}
\begin{figure}[ht!]
    \centering
    \includegraphics[width=0.7\textwidth]{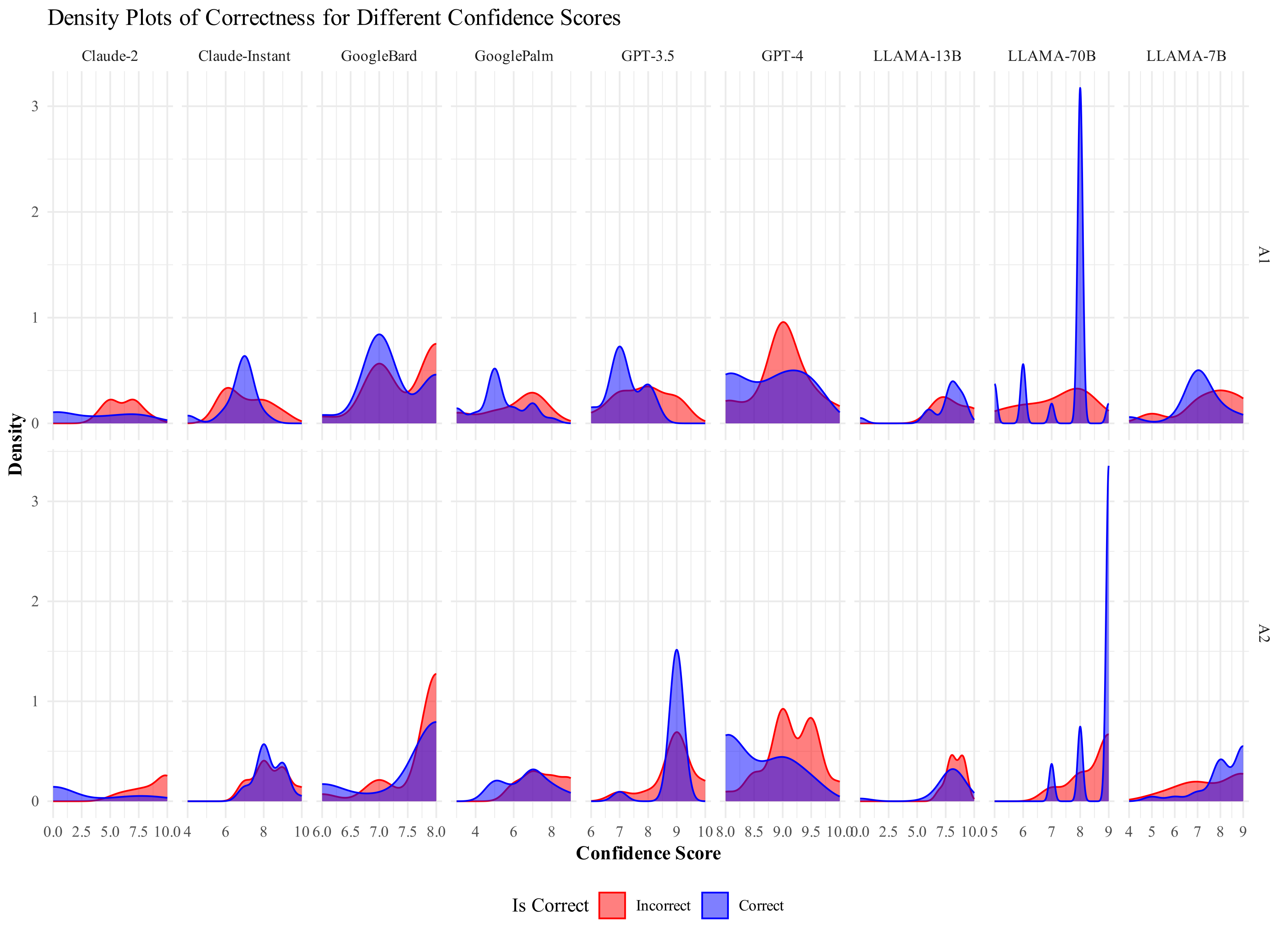}
    \caption{Density Plot of Correctness vs. Confidence Scores for Various Language Learning Models (A1 and A2)}
    \label{fig:dens_corvscon_A}
\end{figure}

The destiny plot in Figure \ref{fig:dens_corvscon_A} represents the relationship between LLMs' confidence score and their correctness in predicting answers. Destiny plot branches each LLM's confidence score into correct and incorrect categories with distinct colors. The higher region in the destiny plot indicates the model is correct or incorrect with specific confidence scores frequently. This plot helps us in providing an initial empirical foundation to access the Dunning-Kruger effect in LLMs. We are interested in where LLMs exhibit high confidence scores and are incorrect or vice versa. This will provide us with information on a misalignment between perceived ability and actual ability.

\begin{figure}[htbp]
    \centering
    \includegraphics[width=0.7\textwidth]{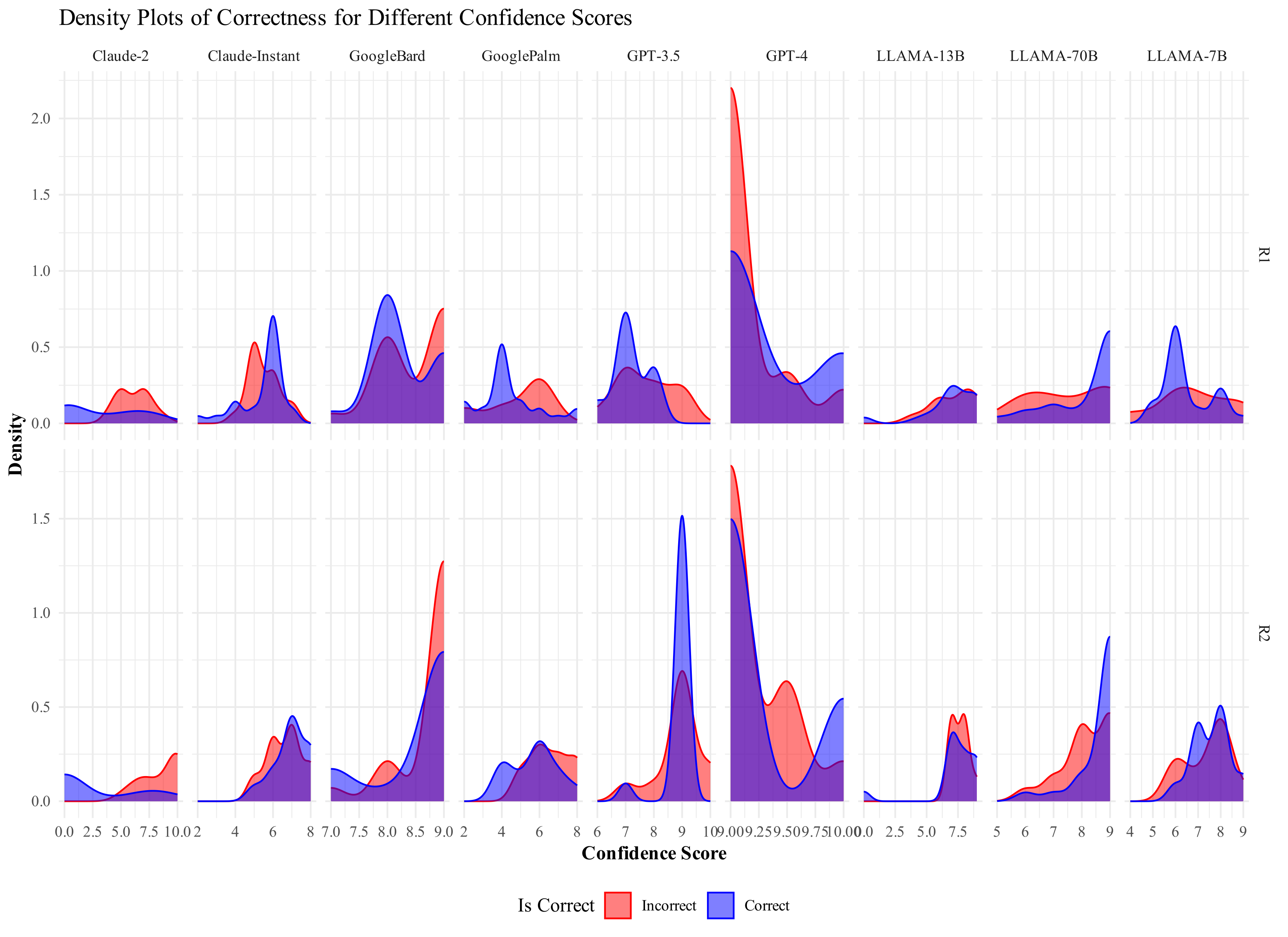}
    \caption{Density Plot of Correctness vs. Confidence Scores for Various Language Learning Models (R1 and R2)}
    \label{fig:dens_corvscon_R}
\end{figure}

\newpage

\subsection{Category Vs Confidence scores}\label{CatVsCon}

\begin{table}[h!]
    \centering
    \caption{Performance and Confidence Metrics of Large Language Models (LLMs) Across Different Categories}
    \label{tab:LLM_Performance}
    \begin{tabular}{llcccc}
        \toprule
        \textbf{Category} & \textbf{LLM} & \textbf{Avg\_A1} & \textbf{Avg\_A2} & \textbf{Avg\_R1} & \textbf{Avg\_R2} \\
        \midrule
        LSAT Reasoning & Claude-2 & 0.80 & 0.00 & 0.60 & 0.00 \\
        LSAT Reasoning & Claude-Instant & 7.00 & 8.20 & 6.00 & 7.20 \\
        LSAT Reasoning & GoogleBard & 7.00 & 7.60 & 8.00 & 8.60 \\
        LSAT Reasoning & GooglePaLM & 5.40 & 6.40 & 4.80 & 5.40 \\
        LSAT Reasoning & GPT-3.5 & 7.00 & 9.00 & 7.00 & 9.00 \\
        LSAT Reasoning & GPT-4 & 8.30 & 8.20 & 9.40 & 9.40 \\
        LSAT Reasoning & LLaMA-13B & 8.20 & 8.10 & 8.40 & 8.40 \\
        LSAT Reasoning & LLaMA-70B & 8.00 & 9.00 & 9.00 & 9.00 \\
        LSAT Reasoning & LLaMA-7B & 7.10 & 8.60 & 6.10 & 7.60 \\
        \midrule
        Mathematical Reasoning & Claude-2 & 5.60 & 4.40 & 5.20 & 4.40 \\
        Mathematical Reasoning & Claude-Instant & 6.80 & 8.90 & 5.50 & 7.20 \\
        Mathematical Reasoning & GoogleBard & 7.40 & 7.80 & 8.40 & 8.80 \\
        Mathematical Reasoning & GooglePaLM & 6.00 & 7.40 & 5.00 & 6.40 \\
        Mathematical Reasoning & GPT-3.5 & 7.60 & 9.00 & 7.60 & 9.00 \\
        Mathematical Reasoning & GPT-4 & 9.20 & 9.20 & 9.40 & 9.40 \\
        Mathematical Reasoning & LLaMA-13B & 8.00 & 8.70 & 7.20 & 8.10 \\
        Mathematical Reasoning & LLaMA-70B & 8.00 & 9.00 & 9.00 & 8.90 \\
        Mathematical Reasoning & LLaMA-7B & 7.80 & 8.40 & 6.80 & 7.40 \\
        \midrule
        Truthful Q\&A & Claude-2 & 6.40 & 8.60 & 6.40 & 8.60 \\
        Truthful Q\&A & Claude-Instant & 6.80 & 8.10 & 5.20 & 6.45 \\
        Truthful Q\&A & GoogleBard & 7.60 & 7.70 & 8.60 & 8.70 \\
        Truthful Q\&A & GooglePaLM & 5.30 & 7.20 & 4.30 & 6.20 \\
        Truthful Q\&A & GPT-3.5 & 7.75 & 8.80 & 7.65 & 8.80 \\
        Truthful Q\&A & GPT-4 & 9.05 & 9.15 & 9.00 & 9.05 \\
        Truthful Q\&A & LLaMA-13B & 7.00 & 7.05 & 6.10 & 6.55 \\
        Truthful Q\&A & LLaMA-70B & 6.70 & 8.20 & 6.90 & 8.00 \\
        Truthful Q\&A & LLaMA-7B & 7.05 & 7.55 & 6.75 & 7.55 \\
        \bottomrule
    \end{tabular}
\end{table}

\newpage

\subsection{ Problem Level Vs Confidence Scores}\label{prolevVsCon}

\begin{table}[h!]
    \centering
    \caption{Average Confidence Scores by Problem Level and LLM}
    \label{tab:problevelvscon}
    \begin{tabular}{cccccccc}
        \toprule
        \textbf{Problem Level} & \textbf{LLM} & \textbf{Avg\_A1} & \textbf{Avg\_A2} & \textbf{Avg\_R1} & \textbf{Avg\_R2} \\
        \midrule
        1 & Claude-2 & 7.250 & 7.125 & 7.000 & 7.125 \\
        1 & Claude-Instant & 7.500 & 9.000 & 6.000 & 7.375 \\
        1 & GoogleBard & 7.500 & 7.250 & 8.500 & 8.250 \\
        1 & GooglePaLM & 6.250 & 7.750 & 5.250 & 6.750 \\
        1 & GPT-3.5 & 8.250 & 9.250 & 8.000 & 9.250 \\
        1 & GPT-4 & 9.250 & 9.125 & 9.500 & 9.250 \\
        1 & LLaMA-13B & 7.750 & 8.000 & 7.500 & 7.750 \\
        1 & LLaMA-70B & 8.000 & 8.625 & 8.500 & 8.250 \\
        1 & LLaMA-7B & 7.625 & 8.250 & 6.875 & 7.500 \\
        2 & Claude-2 & 4.750 & 5.625 & 4.750 & 5.625 \\
        2 & Claude-Instant & 7.000 & 8.250 & 5.750 & 7.000 \\
        2 & GoogleBard & 7.500 & 8.000 & 8.500 & 9.000 \\
        2 & GooglePaLM & 6.000 & 7.000 & 5.500 & 6.000 \\
        2 & GPT-3.5 & 7.500 & 8.125 & 7.500 & 8.125 \\
        2 & GPT-4 & 9.000 & 8.750 & 9.125 & 9.000 \\
        2 & LLaMA-13B & 7.750 & 8.000 & 7.500 & 7.750 \\
        2 & LLaMA-70B & 7.500 & 8.250 & 8.000 & 8.250 \\
        2 & LLaMA-7B & 6.875 & 7.875 & 6.125 & 7.375 \\
        3 & Claude-2 & 4.000 & 5.375 & 4.000 & 5.375 \\
        3 & Claude-Instant & 7.000 & 8.125 & 5.500 & 7.000 \\
        3 & GoogleBard & 7.000 & 7.500 & 8.000 & 8.500 \\
        3 & GooglePaLM & 5.250 & 6.500 & 4.250 & 5.500 \\
        3 & GPT-3.5 & 7.250 & 9.000 & 7.250 & 9.000 \\
        3 & GPT-4 & 8.500 & 8.500 & 9.000 & 9.000 \\
        3 & LLaMA-13B & 5.250 & 6.125 & 4.500 & 5.625 \\
        3 & LLaMA-70B & 6.750 & 8.750 & 7.250 & 8.500 \\
        3 & LLaMA-7B & 6.250 & 7.875 & 5.750 & 7.625 \\
        4 & Claude-2 & 5.250 & 4.875 & 5.000 & 4.875 \\
        4 & Claude-Instant & 7.000 & 8.500 & 5.750 & 7.000 \\
        4 & GoogleBard & 7.500 & 7.750 & 8.500 & 8.750 \\
        4 & GooglePaLM & 6.000 & 7.000 & 5.000 & 6.000 \\
        4 & GPT-3.5 & 7.750 & 9.000 & 7.750 & 9.000 \\
        4 & GPT-4 & 8.875 & 9.250 & 9.125 & 9.375 \\
        4 & LLaMA-13B & 8.750 & 8.000 & 7.750 & 7.875 \\
        4 & LLaMA-70B & 7.250 & 8.750 & 7.750 & 8.500 \\
        4 & LLaMA-7B & 8.000 & 8.000 & 7.500 & 7.625 \\
        5 & Claude-2 & 2.750 & 4.000 & 2.500 & 4.000 \\
        5 & Claude-Instant & 5.750 & 7.750 & 4.375 & 5.750 \\
        5 & GoogleBard & 7.500 & 8.000 & 8.500 & 9.000 \\
        5 & GooglePaLM & 4.000 & 7.000 & 3.000 & 6.000 \\
        5 & GPT-3.5 & 6.875 & 9.125 & 6.875 & 9.125 \\
        5 & GPT-4 & 8.875 & 9.000 & 9.250 & 9.500 \\
        5 & LLaMA-13B & 8.250 & 8.500 & 7.500 & 8.000 \\
        5 & LLaMA-70B & 7.250 & 8.625 & 8.250 & 8.875 \\
        5 & LLaMA-7B & 7.500 & 8.125 & 6.750 & 7.500 \\
        \bottomrule
    \end{tabular}
\end{table}

\end{document}